\documentclass[runningheads]{llncs}
\usepackage{graphicx,wrapfig}
\usepackage{amsmath,amssymb} 
\usepackage{color}
\usepackage{tabu}
\usepackage[width=122mm,left=12mm,paperwidth=146mm,height=193mm,top=12mm,paperheight=217mm]{geometry}
\usepackage{subcaption}
\usepackage{bbm}

\usepackage{hyperref}
\hypersetup{
    colorlinks=true,
}

\newcommand{\etal}{{et al}.\@ }
\DeclareMathOperator{\sign}{sgn}

\newcommand{\argmin}{\mathop{\rm arg~min}\limits}
\newcommand{\VarCustom}{\mathrm{Var}}

\begin{document}
%
%
%
%
%
%

\title{Canonical and Compact Point Cloud Representation for Shape Classification} 

\titlerunning{Canonical and Compact Point Cloud Representation for Shape Classification}
%
\author{Kent Fujiwara\inst{1} \and
Ikuro Sato\inst{1} \and
Mitsuru Ambai\inst{1} \and
Yuichi Yoshida\inst{1} \and
Yoshiaki Sakakura\inst{1} 
}


%
\authorrunning{K. Fujiwara, I. Sato, M. Ambai, Y. Yoshida, and Y. Sakakura}
%

\institute{Denso IT Laboratory, Inc.\\
\email{\{kfujiwara, isato, manbai, yyoshida, ysakakura\}@d-itlab.co.jp}\\
\url{https://www.d-itlab.co.jp/} }

%


\maketitle
\begin{abstract}
      We present a novel compact point cloud representation that is inherently invariant to scale, coordinate change and point permutation. The key idea is to parametrize a distance field around an individual shape into a unique, canonical, and compact vector in an unsupervised manner. We firstly project a distance field to a $4$D canonical space using singular value decomposition. We then train a neural network for each instance to non-linearly embed its distance field into network parameters. We employ a bias-free Extreme Learning Machine (ELM) with ReLU activation units, which has scale-factor commutative property between layers.
We demonstrate the descriptiveness of the instance-wise, shape-embedded network parameters by using them to classify shapes in $3$D datasets. Our learning-based representation requires minimal augmentation and simple neural networks, where previous approaches demand numerous representations to handle coordinate change and point permutation.
\keywords{Point cloud, distance field, extreme learning machines}
\end{abstract}

\begin{figure*}[t]
\begin{center}
   \includegraphics[width=\linewidth]{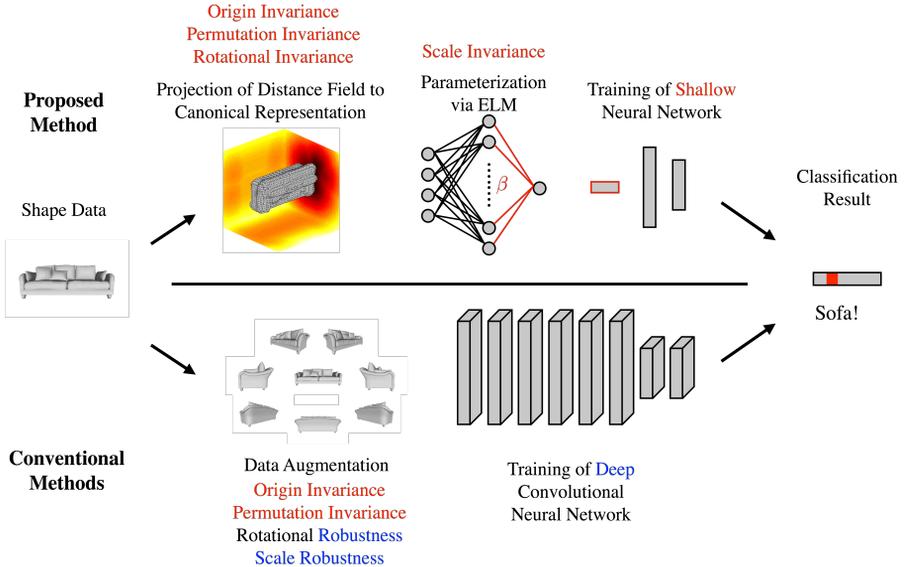}
\end{center}
   \caption{Overview of the proposed method (top) and the conventional methods (bottom). Each step of our method is carefully designed to achieve coordinate invariance, and permutation invariance. The prior work mainly focused on achieving coordinate and point permutation invariance by augmenting the data through random rotation, which are used to train a deep neural network. }
\label{fig:overview}
\end{figure*}

\section{Introduction}
Classifying point cloud data is rapidly becoming a central topic in computer vision, as various sensors are now capable of accurately capturing objects in $3$D. There is a wide range of applications in areas such as autonomous driving and robotics, where accurate geometric information of the environment is critical.

The goal of point cloud classification is to provide an accurate label to a given shape described by a group of unstructured $3$-dimensional points.  There are, however, challenges to overcome to train a model to classify point cloud data accurately. Unlike $2$-dimentional images, points are unordered and can be represented in an arbitrary $3$-dimensional coordinate system. 

These characteristics of unstructured point cloud data require the classifier to acquire the following to achieve minimal generalization error: Invariance to coordinate change, invariance to point permutation, invariance to location of origin, and robustness to resolution. 
More specifically, the output of an ideal point cloud classifier is invariant under arbitrary changes of the origin, orientation, and scale of the $3$D coordinate system used to describe point cloud data.




We propose a novel compact representation that is obtained from a carefully designed unsupervised procedures to achieve invariance to location of origin, scaling, rotation, and point permutation. The proposed method produces a fixed-length vector for each instance, which can be used as a feature for classification.

We take an approach of producing a canonical representation invariant to rotation, scale, location of origin, and point permutation, yet unique to each individual instance. Such invariances can be achieved by aligning permutation-free representation of the shapes to a canonical position without supervision. On the contrary, many of the recently proposed methods~\cite{wu2015,su2015,sinha2016deep} that use deep neural networks attempt to make the model robust to arbitrary $3$D rotation of an object by means of data augmentation technique, where training instances are numerously expanded by applying random rotations around the principle axis, assuming that this information is provided.

We produce a distance field, a scale-covariant implicit function, from an instance of point cloud and project it to a $4$D canonical space. We apply singular value decomposition on the matrix consisting of the coordinates of sampling points in space around the point cloud and the corresponding distance value at that point, from which we obtain the transformation to the canonical space.

We parameterize this canonical representation using a neural network. We train an Extreme Learning Machine (ELM)~\cite{huang2006extreme} with scale-factor commutative property to embed the canonical representation of an individual instance, and produce a representation invariant under scaling and permutation of sampling points. We use the coordinate of the sampling points transformed to the canonical space, and the original distance field information as the input and output of the network, respectively. 

To classify shapes using our representation, we train a separate ELM for each instance and use the parameters of the ELM as a feature of the corresponding instance. We then assign the class label to each feature and conduct supervised training to classify the features accordingly. As our representation is compact, only a shallow neural network is required to obtain very accurate results. 

We demonstrate the validity of our proposal and effectiveness of the resulting compact feature vector through various experiments. We validate the robustness of our representation by comparing our canonical representation to alignment of orientation using principal components. We also conduct experiments using ModelNet 10/40 dataset to demonstrate that our representation requires only subsampling of original data to achieve the accuracy similar to the prior methods that heavily depend on brute-force data augmentation. 


\begin{table*}[t]
\begin{center}
\begin{tabu} to \textwidth {|l|c|c|c|c|c|c|}
\hline
 & \begin{tabular}{@{}c@{}}Rotational \\ Invariant\end{tabular} &  \begin{tabular}{@{}c@{}}Scale \\ Invariant\end{tabular} & \begin{tabular}{@{}c@{}}Origin \\ Invariant\end{tabular} &\begin{tabular}{@{}c@{}}Permutation \\ Invariant \end{tabular} &  \begin{tabular}{@{}c@{}}Robust to \\ Resolution\end{tabular} &  \begin{tabular}{@{}c@{}}Robust to \\ Sampling \end{tabular}  \\
\hline
Voxel-based~\cite{wu2015} & $-$ & $-$ & \checkmark&\checkmark & $-$ & $-$ \\
Image-based~\cite{su2015} & $-$ & $-$ & \checkmark&\checkmark & \checkmark &$-$\\
Intrinsic~\cite{sinha2016deep} & $-$ &\checkmark& \checkmark & \checkmark &  \checkmark&\checkmark \\
Point-based~\cite{qi2016pointnet} & $\bigtriangleup$ & $\bigtriangleup$ & \checkmark& \checkmark & \checkmark &\checkmark\\
Our method & \checkmark & \checkmark & \checkmark& \checkmark & \checkmark  &\checkmark \\
\hline
\end{tabu}
\end{center}
\caption{Contributions of prior methods and our method. The marks represent: Good (\checkmark), fair ($\bigtriangleup$), and poor ($-$). }
\label{tab:contribution}
\end{table*}

\section{Related Work}

Classification of $3$-dimensional data is an active area of research, as it has the potential to be utilized in various applications, such as tasks in robotics, autonomous driving and surveillance. Most early works focus on defining a robust representation that can be used to retrieve certain shapes from database of $3$D models. We refer the readers to a survey~\cite{tangelder2004} for a comprehensive overview. 

With the advances in deep learning techniques and more datasets~\cite{song2015,chang2015} becoming available, classifying shapes from point clouds is quickly becoming a major task in computer vision. Most methods focus on finding a ''convolution-friendly'' representation that can be used as input to convolutional neural networks. Common approaches are voxel-based, image rendering-based, cross-section-based, and point-based methods. Table~\ref{tab:contribution} shows the common approaches and the invariance and robustness they achieve. 

Voxel-based methods~\cite{wu2015,maturana2015,wu2016learning,sedaghat2017orientation} convert point clouds into voxel data and feeds them to a convolutional neural network to classify them according to the provided labels. Convolution is conducted in a similar manner to CNN-based $2$D image classification methods, with $3$D filters to convolve local voxels. These methods avoids the issue of point permutation by embedding points into voxels. As the representation cannot handle coordinate change, data augmentation is conducted by rotating the original shape and voxelizing it from another viewpoint. Our method is similar in that we employ a spatial representation of the shapes. However, our representation is invariant to coordinate change, getting rid of the necessity to augment data by random rotations. Our method is also robust to resolution, while the accuracy of voxel-based methods depend heavily on the size of the voxels.

Image-rendering-based methods~\cite{su2015,Wang2017} attempt to use the rendered images to classify the depicted shapes. Instead of using $3$D representation, these methods convert them into $2$D images and apply CNN that is typically used for image classification tasks. This approach avoids the point permutation issue by projecting them onto the image plane. The images are covariant to coordinate change, therefore, the target shapes are rendered from multiple viewpoints. The accuracy of these methods depend on the number of viewpoints. Our method, in contrast, projects the $3$D data into $4$D canonical shape to avoid data augmentation. There are also attempts to combine the volumetric and the image data~\cite{qi2016volumetric} to improve the accuracy of classification.

Another class of methods attempt to convert the shapes into a common representation. Sinha \etal~\cite{sinha2016deep} creates an authalic mapping of the $3$D models, which can be fed into a $2$D CNN for classification. To handle coordinate change on the $2$D plane, they conduct data augmentation by rotating the authalic maps. Shi~\etal~\cite{shi2015deeppano} projects the shape onto a cylinder, which is dissected into a $2$D image to be used for CNN classification. These methods requires an axis parallel to the direction of gravity. There are shapes whose gravitational direction is difficult to determine. Our canonical representation does not require such supervision, as we achieve rotational invariance by solving for projections to the canonical space. 

\begin{figure}
\begin{subfigure}[t]{0.45\textwidth}
        \centering
        \includegraphics[width=0.9\linewidth]{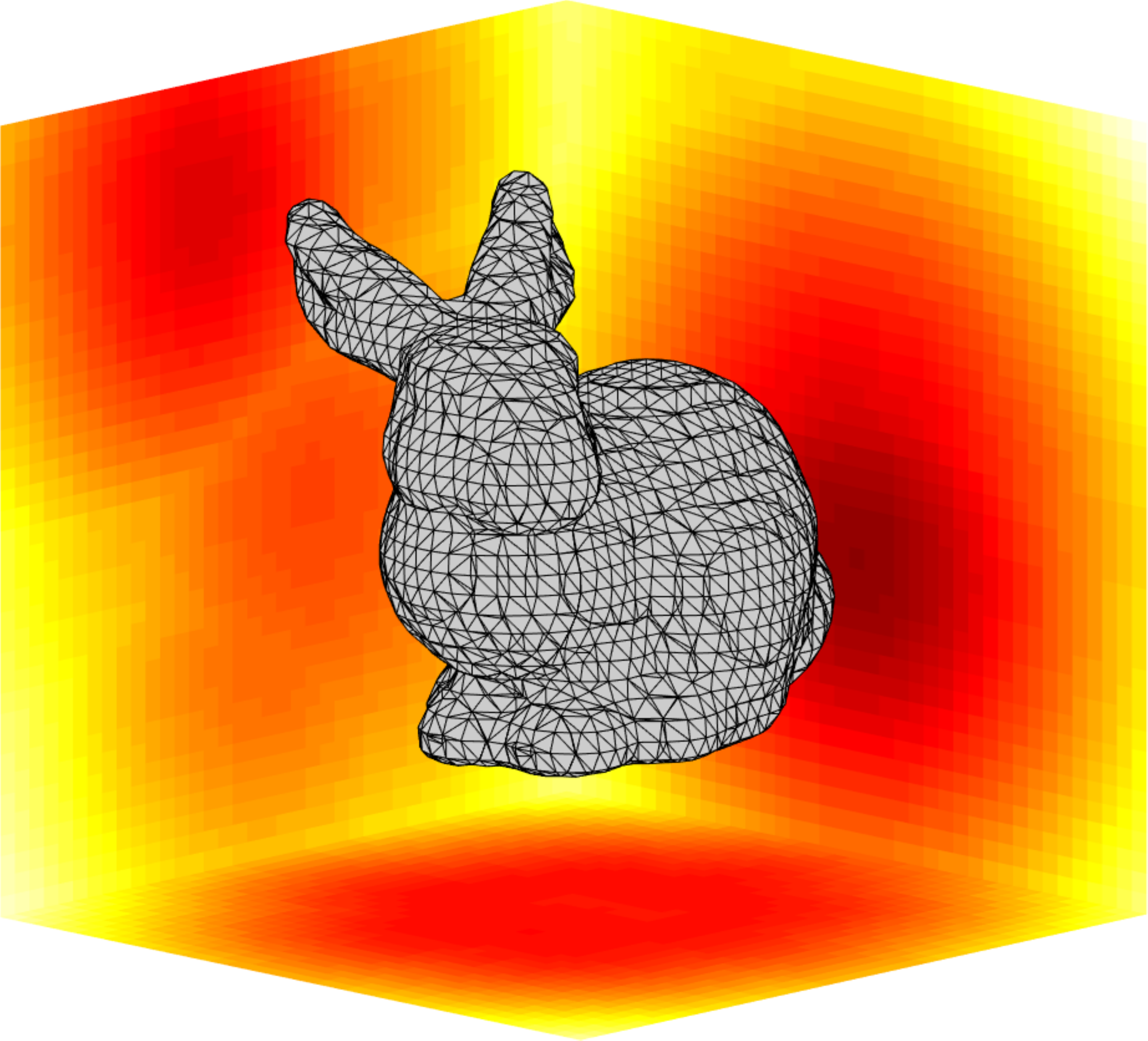}
        \caption{}
        \label{fig:bun}
\end{subfigure}%
\begin{subfigure}[t]{0.55\textwidth}
        \centering
        \includegraphics[width=0.9\linewidth]{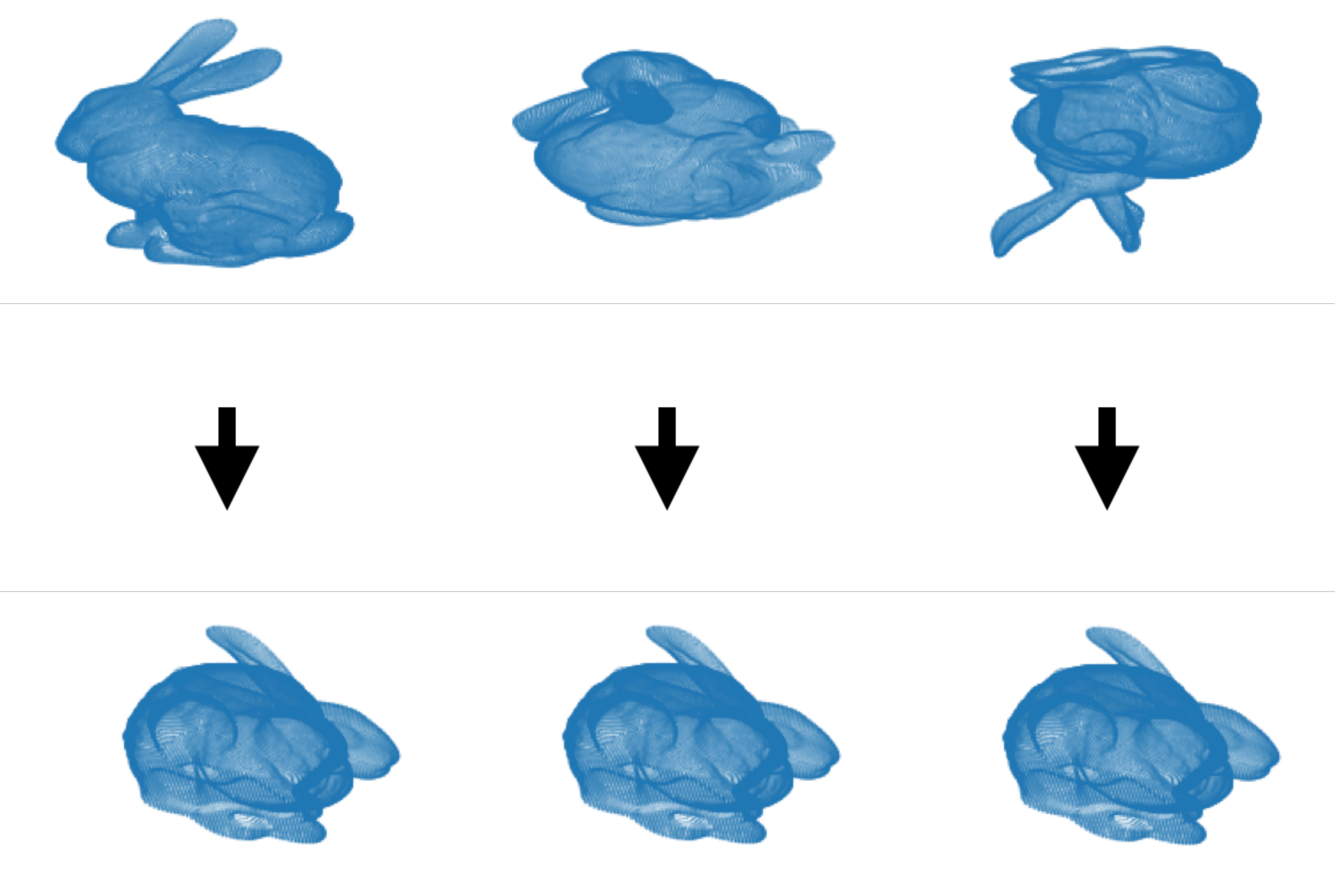}
        \caption{}
        \label{fig:canonical}
\end{subfigure}%
   \caption{(a) Distance field of Stanford bunny. Regions with low values (red) are closer to the target surface, and those with high values (yellow) are far. (b) Rotation invariant representation. For the purpose of visualization, the original point cloud, which can be considered as the $0$-level set of the distance field, is transformed according to the transformation $\bar{\mathbf{V}}$ obtained from SVD.  Top row: Stanford bunny (left) rotated at random angles are shown in the right. Bottom row: The transformed positions of the surface. Despite randomly rotating the original model, our method aligns the distance fields to a unique position.}
\label{fig:canonoverall}
\end{figure}

There are attempts to process point information directly. Qi~\etal~\cite{qi2016pointnet,qi2017pointnet++} feeds the point cloud data into a neural network directly. The neural network of the method includes affine transformation layers that align shapes to avoid coordinate ambiguity and facilitate the final classification. The method avoids the issue of point permutation by pooling the points in higher dimensional space and use it as the feature. Our approach differs in the fact that we project the implicit t representation of point clouds to a canonical space, which is then parameterized into a fixed length vector. Our representation also achieves scale invariance with the combination of the distance field and scale-factor commutative neural network.  Klokov and Lempitsky~\cite{klokov2017escape} converts the unstructured points into k-d trees, making it invariant to coordinate change. The tree representation is used to train a neural network to classify point cloud data. The structure may differ depending on the density of the point data, an issue we overcome by implicitly representing the shapes. Depth map is also used to classify shapes~\cite{soltani2017synthesizing,zanuttigh2017deep}.


\section{Proposed Method}

We propose a novel method that overcomes the difficulty of unstructured point cloud representation by achieving invariance to coordinate change and point permutation. Our method to convert point cloud data into canonical representation consists of three steps: Calculation of distance field from point cloud data, canonical projection of the distance field, and network embedding of the canonical representation. 


\subsection{Implicit Representation: Distance Field}

There are two key reasons behind why we choose to represent point cloud data using an implicit representation, the distance field. 

The first is that the distance field achieves invariance to point permutation. The distance field embeds the distance between points and the space around them, therefore the same set of points results in the same distance field, regardless of the ordering of points. This solves one of the first issues in point cloud representation. There are other ways to achieve the invariance. Volumetric occupancy, as employed in voxel-based methods~\cite{wu2015} is also invariant to point permutation.

The second reason, which is key to our method, is that distance fields are covariant to scale. When the coordinate values are scaled, the distance is also scaled by the same factor. This property of distance field is critical in our method, where we embed the representation to a network with scale-factor commutative property, thus achieving scale invariance. The volume occupancy representation is not covariant to scale, therefore cannot achieve scale invariance as in our method.



Distance field is an implicit representation of a shape that captures the distance between a shape and the space around it. Fig.~\ref{fig:bun} is a distance field calculated from the Stanford bunny. Given a shape $\mathcal{P}$ consisting of $n$ points $\mathbf{p} \in \mathbb{R}^3$,  the distance function $\phi$ at a point in the surrounding space $\mathbf{x} \in \mathcal{X}$ is defined as 
\begin{equation}
\phi(\mathbf{x}) = \min\limits_{\mathbf{p} \in \mathcal{P}} \| \mathbf{x} - \mathbf{p}\|\,.
\label{eq:distancefield}
\end{equation}
In practice, we sample $m$ points from inside a sphere around the target shape. We then measure the distance from these sampling points to the nearest point on the shape surface to construct a distance field. Similar to conventional methods, we achieve origin invariance by aligning the centroid of the model and the sphere.

The distance field representation, as is, is variant under coordinate change. When the sampling sphere and the target shape is rotated, the coordinates of the sampling points change while the distance field maintains its values. 

\subsection{Projection to Canonical Space}

We propose to project the distance field onto a $4$-dimensional canonical space to achieve invariance to rotation. We introduce a sampling data matrix $\mathbf{X} =  \begin{bmatrix} \mathbf{x}_1, & \mathbf{x}_2, & \cdots &, \mathbf{x}_m\end{bmatrix}^{\top} \in \mathbb{R}^{m \times 3}$  and the corresponding sampled distance vector $\mathbf{\Phi}_{\mathcal{P}}(\mathbf{X}) =  \begin{bmatrix} \phi(\mathbf{x}_1), & \phi(\mathbf{x}_2), & \cdots &, \phi(\mathbf{x}_m)\end{bmatrix}^{\top} \in \mathbb{R}^{m}$ and concatenate them into a matrix $\mathbf{M} = \begin{bmatrix} \mathbf{X} &  & \mathbf{\Phi}_{\mathcal{P}}(\mathbf{X})\end{bmatrix}$. We then apply singular value decomposition (SVD) on $\mathbf{M}$ to obtain 

\begin{equation}
\mathbf{M} = \mathbf{U}{\mathbf{S}}\mathbf{V}^{\top}\,.
\label{eq:svd1}
\end{equation}

Interestingly, $\mathbf{U}{\mathbf{S}}$ is independent to an arbitrary rotation of the coordinate system. If the coordinates are rotated by an arbitrary rotation matrix $\mathbf{R}$, $\mathbf{M}$ becomes $\mathbf{M}' = \begin{bmatrix} \mathbf{X}\mathbf{R} &  & \mathbf{\Phi}_{\mathcal{P}}(\mathbf{X})\end{bmatrix}$. $\mathbf{M}$ and $\mathbf{M}'$ is related by $\mathbf{M}' = \mathbf{M}\begin{bmatrix} \mathbf{R} & \mathbf{0} \\ \mathbf{0} & 1\end{bmatrix}$. Since the SVD of $\mathbf{M}$ is as shown in eq.~(\ref{eq:svd1}), SVD of $\mathbf{M}'$ is
\begin{equation}
\mathbf{M}' = \mathbf{U}{\mathbf{S}} \mathbf{V}^{\top}\begin{bmatrix} \mathbf{R} & \mathbf{0} \\ \mathbf{0} & 1\end{bmatrix}\,.
\label{eq:svd2}
\end{equation}
This indicates that $\mathbf{U}{\mathbf{S}}$ is independent of the arbitrary rotation  $\mathbf{R}$.

As the results from SVD may contain ambiguity of sign, we propose to fix the signs of the results through the following process. As $\mathbf{S}$, the singular values, are all positive, the task is to fix the sign ambiguity between $\mathbf{U}$ and $\mathbf{V}^{\top}$. As $\mathbf{U}$ is orthogonal, we can apply the following to obtain

\begin{equation}
\mathbf{U}^{\top} \mathbf{M}= \mathbf{S} \mathbf{V}^{\top}\,.
\label{eq:transpose}
\end{equation}
It is now clear that both sides of eq. (\ref{eq:transpose}) must have the same sign. We fix the sign in a way that is specific to each individual. Therefore, we focus on the last column of $\mathbf{M}$, which contains the distance values $\mathbf{\Phi}_{\mathcal{P}}(\mathbf{X})$ and determine the sign so that $\mathbf{U}^{\top}\mathbf{\Phi}_{\mathcal{P}}(\mathbf{X}) \geq 0$. Therefore, given an input data, we obtain the sign of $\mathbf{U}$ and $\mathbf{V}$ by calculating $\mathbf{c} = \sign (\mathbf{U}^{\top}\mathbf{\Phi}_{\mathcal{P}}(\mathbf{X}))$ and applying the signs to $\mathbf{V}^{\top}$ to retrieve 
 \begin{equation}
 \bar{\mathbf{V}}^{\top} = \mathbf{V}^{\top} \mathbf{C}\,,
 \label{eq:updateV}
 \end{equation}
where $\mathbf{C}$ is a diagonal matrix that contains $\mathbf{c}$ in its diagonal elements.

The matrix $\mathbf{M}$ can be considered as a $4$-dimensional data that consists of sampling points scattered within a sphere and the value from the distance function. As the distribution of the first three dimensions are most likely uniform, SVD transforms the matrix based on the distance values, which is unique to each instance. Therefore, $\bar{\mathbf{V}}$ transforms $\mathbf{M}$ to a $4$-dimensional canonical space, in which the shape is uniquely aligned regardless of the initial pose. We visualize the result by treating the points of the Stanford Bunny as the sampling points $\mathbf{X}$ and setting the corresponding distance values $\mathbf{\Phi}_{\mathcal{P}}(\mathbf{X})$ to $0$ to represent the $0$-level set. As the last singular value becomes zeros, we can observe the transformation result by depicting the first 3 columns of $\bar{\mathbf{M}} = \mathbf{M}\bar{\mathbf{V}}$. Fig.~\ref{fig:canonical} shows that regardless of the initial pose of the Bunny, the models are all aligned to a unique pose based on the distribution of the distance field. We will demonstrate the robustness to surface point density in the experimental section.


The canonical representation is now invariant to rotation, as a shape in various poses is aligned in the canonical space. However, it remains covariant to scale change, as the distance values change accordingly to its scale. Also, the representation still has ambiguity regarding permutation of sampling points.


\subsection{Parameterization of Canonical Representation}

We train a neural network to embed our canonical point cloud representation into parameters in an instance-wise fashion. The non-parametric representation of the canonical point cloud is parameterized through network training.

Typical multi-layer neural networks have extremely large possibility of weight combinations, as weights at different layers are simultaneously optimized. Due to this characteristic, 
weights are not uniquely determined for a given instance. 
We, therefore, choose a specific type of a neural network, namely an Extreme Learning Machine (ELM)~\cite{huang2006extreme} to embed the canonical representation. 

An ELM is a two-layer feedforward neural network whose weights are set at random. To define the input of ELM, we decompose $\bar{\mathbf{M}}$ into 
\begin{equation}
\bar{\mathbf{M}} = \mathbf{M} \bar{\mathbf{V}} = \mathbf{X}\bar{\mathbf{V}}_{1:3,:} +   \mathbf{\Phi}_{\mathcal{P}}(\mathbf{X})\bar{\mathbf{V}}_{4,:}\,,
\label{eq:decompose}
\end{equation}
where $\bar{\mathbf{V}}_{1:3,:}$ is the first three rows of $\bar{\mathbf{V}}$ (or first three columns of $\bar{\mathbf{V}}^{\top}$) and $\bar{\mathbf{V}}_{4,:}$ is the last column of  $\bar{\mathbf{V}}$. As we will use  $\mathbf{\Phi}_{\mathcal{P}}(\mathbf{X})$ as the output of the ELM, we get rid of the second term on the right hand side of eq. (\ref{eq:decompose}) to avoid trivial solution. Thus the input to the ELM is
\begin{equation}
\bar{\mathbf{X}} = \mathbf{X}\bar{\mathbf{V}}_{1:3,:}\,,
\end{equation} 
We train the ELM so that $\bar{\mathbf{x}}_i \in \mathbb{R}^{4}$ is ideally converted to $\phi_{\mathcal{P}}(\mathbf{x}_i) \in \mathbb{R}$ for all $i \in \{1, 2, \cdots, m\}$, where $\bar{\mathbf{x}}_i$ is the $i$-th row of $\bar{\mathbf{X}}$. 

\begin{figure}[t]
\begin{subfigure}[t]{0.5\textwidth}
        \centering
        \includegraphics[width=0.9\linewidth]{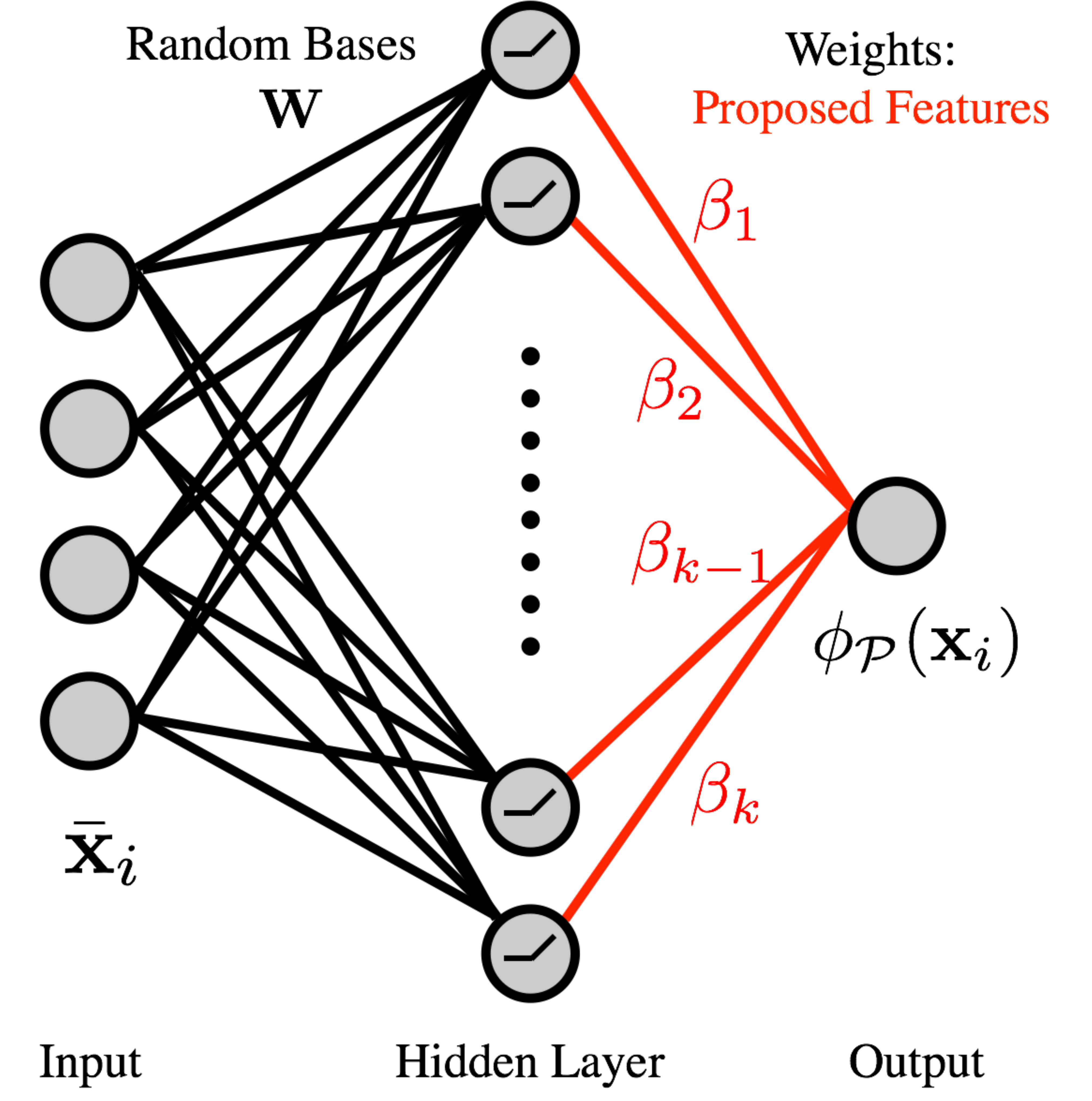}
        \caption{}
        \label{fig:elm}
\end{subfigure}%
\begin{subfigure}[t]{0.5\textwidth}
        \centering
        \includegraphics[width=\linewidth]{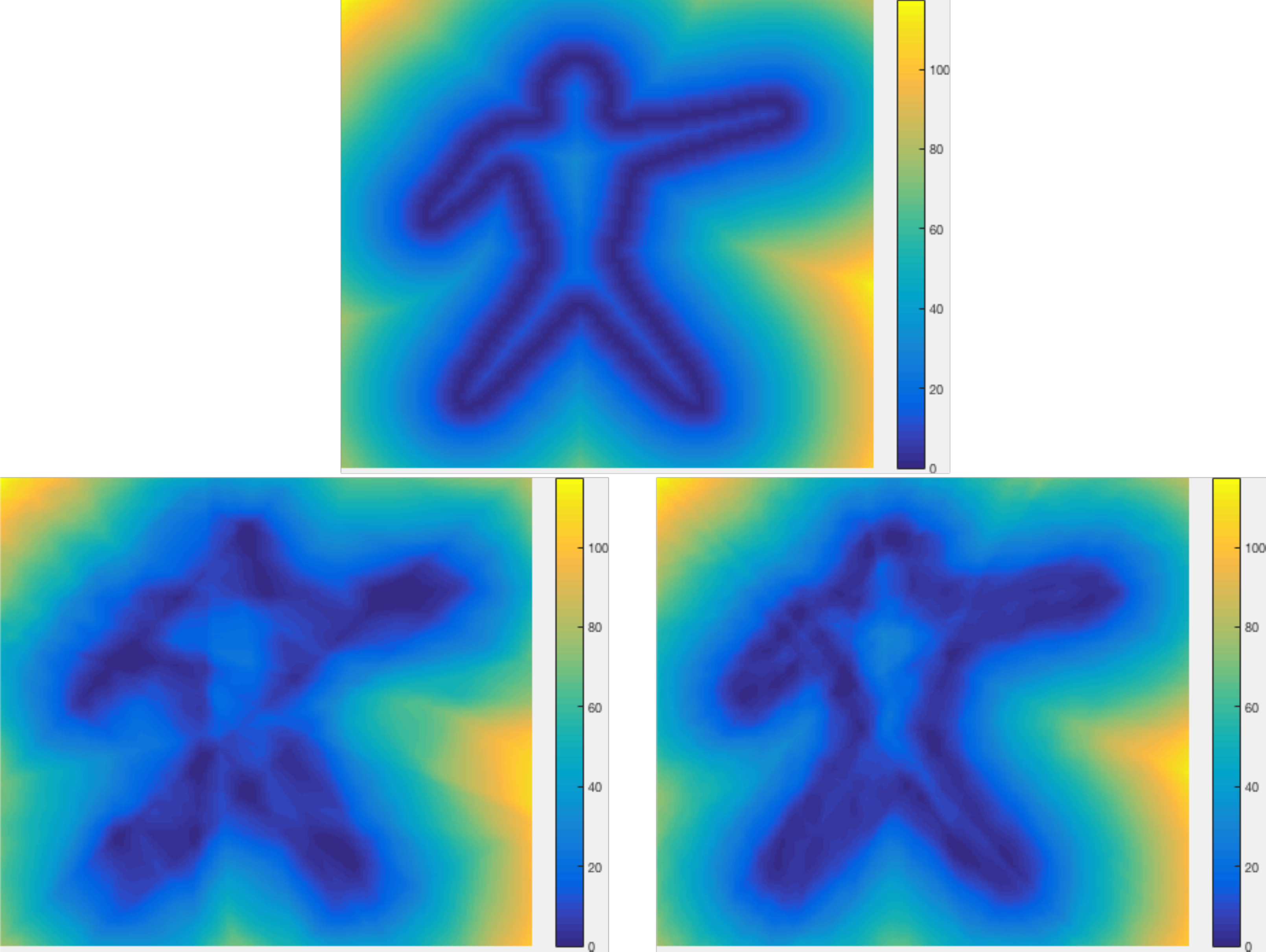}
        \caption{}
        \label{fig:dt2d}
\end{subfigure}%
   \caption{(a) We train the ELM to learn the relationship between the input, sampling points in the canonical coordinates $\bar{\mathbf{x}} \in \bar{\mathbf{X}}$ and the output, the distance value in the original coordinate system. We use the obtained parameters, shown in red, as the feature of the instance. In our proposed method, one ELM is trained to embed a single instance. (b) Distance field of a figure and the reconstruction results from ELM. Top: Ground truth. Bottom Left: Reconstruction from ELM with 300 nodes. Right: Reconstruction from ELM with 1000 nodes.}
\label{fig:overallelm}
\end{figure}

Generally, a bias column is added to the input to aid the learning process. The input is now enhanced to $\tilde{\mathbf{X}} = \begin{bmatrix}\bar{\mathbf{X}} & \mathbf{b}\end{bmatrix} \in \mathbb{R}^{m \times (4 + 1)}$, where $\mathbf{b}$ is a bias column, generally set as $\mathbbm{1}$.

We formulate the objective function of ELM as
\begin{equation}
\boldsymbol{\beta}^{*} = \argmin_{\boldsymbol{\beta}} \| \mathbf{\Phi}_{\mathcal{P}}(\mathbf{X}) - \boldsymbol{\beta}^{\top} f (\mathbf{W}\tilde{\mathbf{X}}^{\top})\|_{F}^{2}\,,
\label{eq:elm}
\end{equation}
where $\boldsymbol{\beta} \in \mathbb{R}^{k}$ is the network parameters, $f$ is a non-linear activation function, $\mathbf{W} \in \mathbb{R}^{k \times (4+1)}$ is the random weight. To obtain the parameters $\boldsymbol{\beta}$ so that the network output matches the target $\mathbf{\Phi}_{\mathcal{P}}(\mathbf{X})$, we simply need to solve for the pseudo-inverse of  $\mathbf{H}  = f (\mathbf{W}\tilde{\mathbf{X}}^{\top})$ to obtain $\boldsymbol{\beta}^{*} = \mathbf{H}^{\dagger}\mathbf{\Phi}_{\mathcal{P}}(\mathbf{X})$, or more robustly, obtain
\begin{equation}
\boldsymbol{\beta}^{*}  = (c\mathbf{I} + \mathbf{H}^{\top} \mathbf{H})^{-1} \mathbf{H}^{\top} \mathbf{\Phi}_{\mathcal{P}}(\mathbf{X})\,,
\end{equation}
where, $c$ is a constant added to the diagonal elements of $ \mathbf{H}^{\top} \mathbf{H}$.

As $\mathbf{W}$ of ELM are fixed to a random value, the parameters $\boldsymbol{\beta}$ corresponding to them are uniquely determined. We will exploit this characteristic of ELM to provide a unique set of weights of fixed length $k$ for each instance of point cloud data. 

By embedding the canonical representation into an ELM, we are able to avoid dealing with permutation of the sampling points. Regardless of the order of the input data, the same canonical representation can be obtained from the ELM.

To intuitively understand the ELM's ability to capture the distance field representation, we embedded a $2$-D distance field for the sake of visualization. Fig.~\ref{fig:dt2d} shows the true distance field calculated from a contour of a person, and its reconstruction from different ELM setups. As can be observed from the reconstruction, the number of ELM nodes affects the accuracy of the reconstruction. Note that details are captured when the number of nodes increases.

Furthermore, we achieve scale invariance by configuring our ELM settings. We employ the rectified linear unit (ReLU)~\cite{nair2010rectified} as the activation function $f$ 
and set $c = \VarCustom(\bar{\mathbf{X}})$ and $\mathbf{b} = \sigma(\bar{\mathbf{X}}) \mathbbm{1}$, where $\VarCustom(\bar{\mathbf{X}})$ and $\sigma(\bar{\mathbf{X}})$ are the variance and the standard deviation of all the $4m$ elements in $\bar{\mathbf{X}}$. 
%


To demonstrate the scale invariance, we consider the case of applying a scaling factor $s$ to the original data $\bar{\mathbf{X}}$. This means that $\sigma(s\bar{\mathbf{X}}) = s \sigma(\bar{\mathbf{X}})$, as $\sqrt{\VarCustom(s\bar{\mathbf{X}})} = \sqrt{s^2\VarCustom(\bar{\mathbf{X}})} = s\sigma(\bar{\mathbf{X}})$. Therefore, when $\bar{\mathbf{X}}$ is scaled by $s$,  $\begin{bmatrix} s\bar{\mathbf{X}} & s\mathbf{b}\end{bmatrix} = s\tilde{\mathbf{X}}$. The output $\mathbf{\Phi}_{\mathcal{P}}(\mathbf{X})$, the distance value, is also covariant to scale change. Therefore, the relationship between the input and the output of ELM can be written as 

\begin{equation}
s\mathbf{\Phi}_{\mathcal{P}}(\mathbf{X}) \approx \boldsymbol{\beta}^{*\top} f(\mathbf{W}s\tilde{\mathbf{X}})\,.
\label{eq:elmscale}
\end{equation}

As ReLU lets through positive values as is, the scale element can be moved outside the activation function, leading to
\begin{equation}
s\mathbf{\Phi}_{\mathcal{P}}(\mathbf{X}) \approx s\boldsymbol{\beta}^{*\top} f(\mathbf{W}\tilde{\mathbf{X}})\,.
\label{eq:elmscaleafter}
\end{equation}
The scaling element cancels each other out, leaving the network parameters $\boldsymbol{\beta}$ unchanged. This allows our modified version of ELM to be invariant to scaling of the target models. Note that variants of ReLU, such as LeakyReLU~\cite{maas2013rectifier}, also preserves this quality, and thus can be used as the activation function.

\subsection{Shape Classification with Proposed Representation}

The proposed method above encapsulates the distance field of each individual instance in a fixed number of parameters $\boldsymbol{\beta}^{*}$. Note that one ELM represents one individual instance. To represent multiple shapes, we train a different ELM separately, so that one ELM corresponds to one instance. 

As shapes are now represented by a compact set of parameters, we can train a classifier to classify shapes according to provided labels. Since we have successfully encapsulated the objects in a small number of parameters, only a shallow neural network is required to accurately distinguish object classes. We will demonstrate the powerfulness of our representation in the following section.




\section{Experiments}

\subsection{Robustness of Representation}
We firstly evaluate the robustness of the proposed canonical projection. We compared our projection approach with another commonly used approach to align shapes, which is by conducing principal component analysis (PCA) on surface points themselves. We used the Stanford bunny~\cite{turk1994zippered}, which consists of approximately $30000$ vertices, and randomly selected  $10000, 5000,$ and $1000$ points from the point cloud $10$ different times. 

For PCA based alignment, we conducted PCA on each of the sampled point sets directly, and computed  the eigenvector corresponding to the largest eigenvalue. We then measured the cosine similarity between all the pairs of the first eigenvectors. 
For our method, we obtained the distance field by measuring the distance from random sampling points in a unit sphere to the selected object points. We conducted SVD on the matrix created by the sampling point coordinates and its distance function output. We measured the cosine similarity of the first column of $\bar{\mathbf{V}}$, the projection to the first axis of the canonical space. 

Table~\ref{tab:svdacc} shows the mean and the standard deviation of all the cosine similarity values in each method. PCA directly on surface points returns a relatively stable transformation at highly sampled data, but quickly deteriorates when the number of subsampled points is small. Our method, in contrast, is robust to both the surface point sampling density and the spatial sampling density, as both the mean and the standard deviation values barely change. This suggests that our alignment method, regardless of the density of the point cloud of the target shape, produces a robust representation. This is due to the fact that the distance field is a non-parametric implicit representation of the shape and is unaffected by the resolution of the original point cloud.

\begin{table}[t]
\begin{center}
\begin{tabular}{|l|c|c|c|c|}
\hline
methods  & 10000 pts & 5000 pts & 1000 pts\\
\hline
PCA mean& 0.9997 &  0.9384 & 0.7549\\
PCA st. d.& 0.00027 &  0.1107 & 0.22898\\
\hline
ours 50000 smp. $\mu$& 0.9999 &  0.9999 & 0.9970\\
ours 50000 smp. $\sigma$& 4.43E-07 &  9.73E-07 & 0.0057\\
\hline
ours 10000 smp. $\mu$& 0.9999 &  0.9999 & 0.9997\\
ours 10000 smp. $\sigma$& 2.76E-06 &  4.66E-06 & 0.0003\\
\hline
ours 5000 smp. $\mu$& 0.9999 &  0.9999 & 0.9995\\
ours 5000 smp. $\sigma$& 1.05E-06 &  1.94E-05 & 0.0005\\
\hline
\end{tabular}
\end{center}
\caption{Comparison of the deviation of principle axes. We compared the average pairwise cosine similarity and the standard deviation of the vector transforming the input to the first principle axis. }
\label{tab:svdacc}
\end{table}

When the target object is highly symmetrical,
the variance of the distance field will be identical in all directions within the sampling space. Therefore, the SVD may return various transformations, as there are infinite possible principal axes. However, since the distance fields of symmetrical shapes are also symmetrical, its effect on the classification accuracy is expected to be minimal.


\subsection{3D Shape Classification with Single Representation}

We then demonstrate the effectiveness of the features obtained from our method using the benchmark dataset Modelnet 10/40~\cite{wu2015}. We first normalized the CAD models in the dataset by setting it to zero-mean, encapsulating the models in a unit sphere, and then uniformly sampling over the model surface, as conducted in most methods such as ~\cite{qi2016pointnet}. We then set random sampling points within the unit sphere and calculated the distance field for all of the shapes in the training and testing data. 
We trained one ELM per each sample in the dataset, and converted all the data in the training and testing set to ELM parameters, which are used as features, each of which represents an individual shape. The random bases are orthogonalized, as conducted in \cite{tang2016extreme}, to improve the result of regression. Note that the random bases are identical among the ELMs, as all the shapes have to be represented in the same parametric space. 

\subsubsection{Effects of Elements in the Proposed Method}

We first analyze the effect of various elements in the proposed method. To observe the effects of the number of sampling points, the number of ELM nodes, and the layers of the classification neural network. For sampling points, we prepared $2048, 4096, 8192$ random sampling points, and for ELM nodes we observed results for $16, 32, 64, 128, 256, 512, 1024$ nodes.  $2048$ surface points are sampled from the original $3$-D models.

For classification, we prepared a base network consisting of $3$ hidden layers with $512, 256, 128$ nodes. We then provided additional layers consisting of $512$ nodes to the network to observe the change in the classification accuracy. For the purpose of observation, no dropout has been set between the layers.

\begin{figure}[t]

\captionsetup[subfigure]{justification=centering}
\centering
\subcaptionbox{Base network 1 \\ 512-256-128 \label{fig:10-1}}{\includegraphics[width=.32\linewidth]{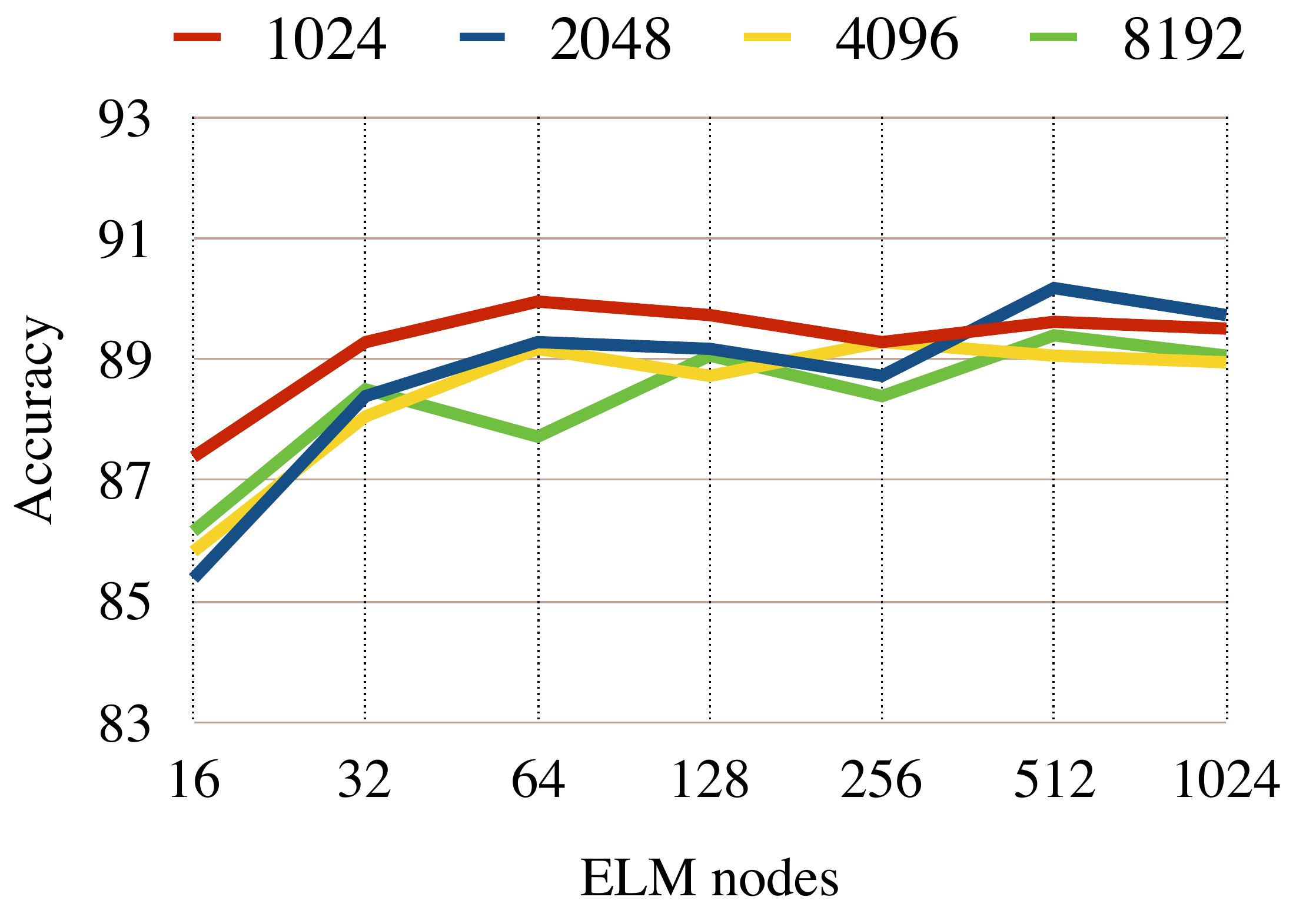}}
\centering
\subcaptionbox{Network 2  \\ 512-512-256-128 \label{fig:10-2}}{\includegraphics[width=.32\linewidth]{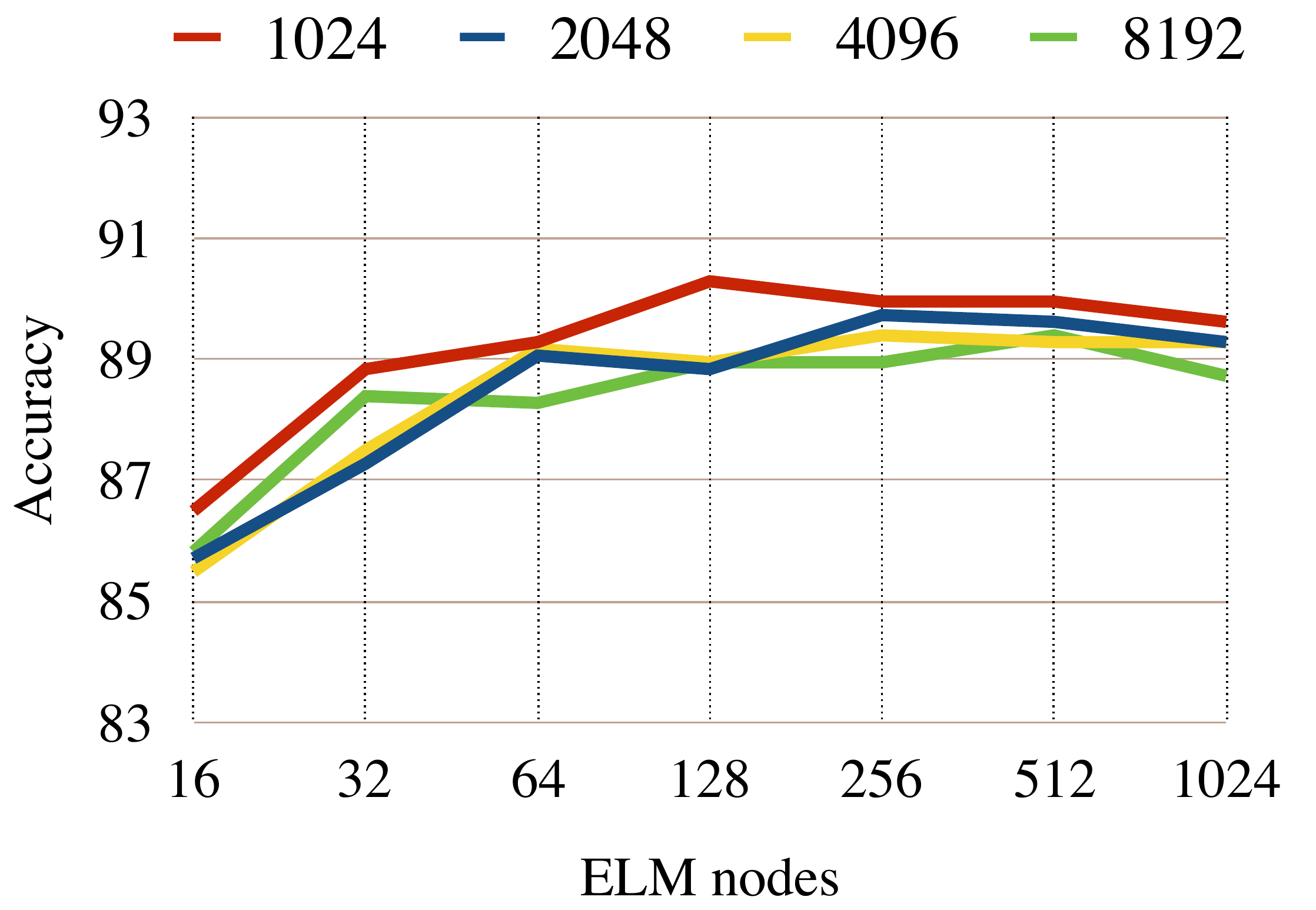}}
\centering
\subcaptionbox{Network 3 \\  512-512-512-256-128 \label{fig:10-3}}{\includegraphics[width=.32\linewidth]{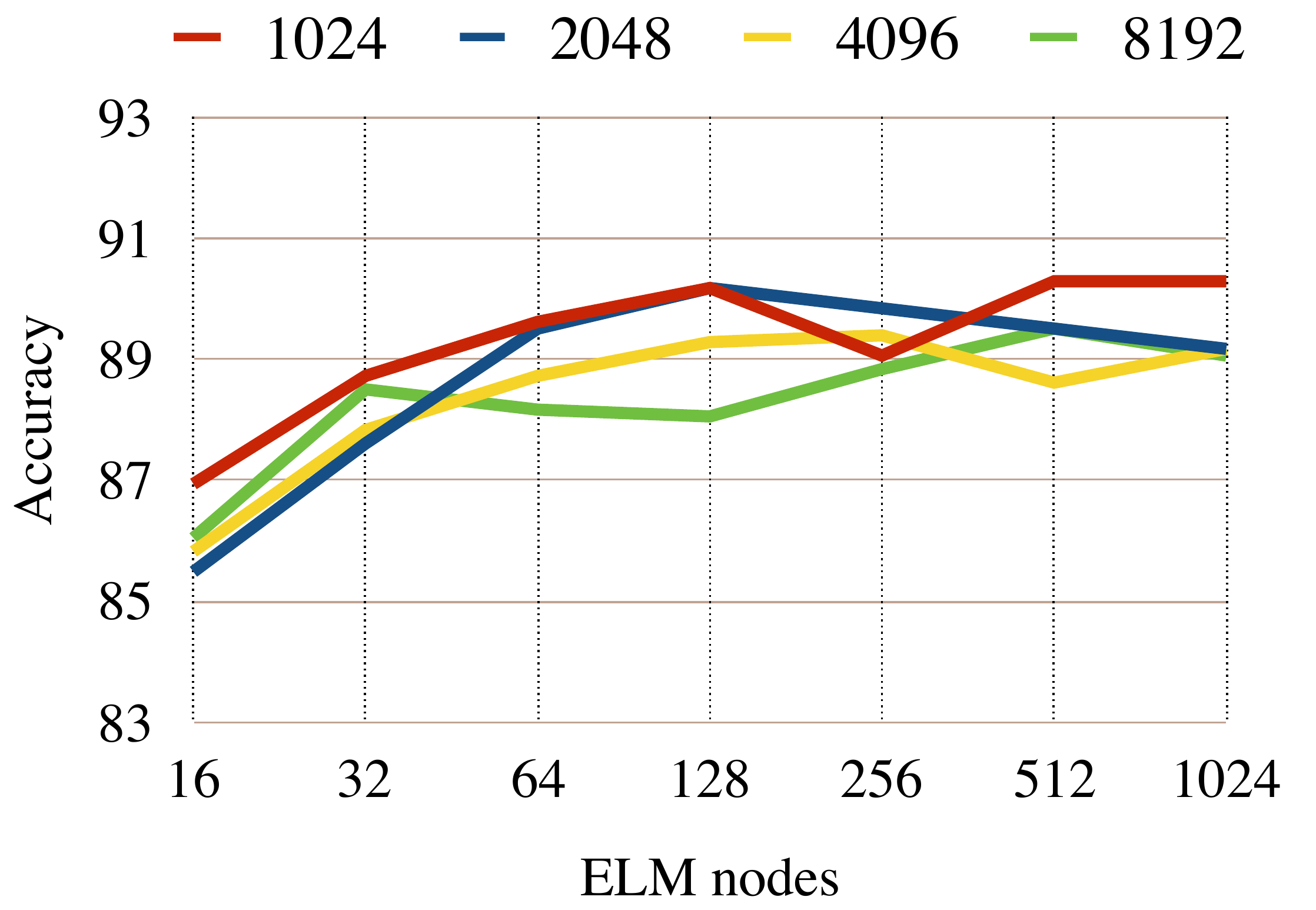}}

   \caption{Accuracy of the proposed method for ModelNet 10 at different sampling densities. The numbers beside the lines indicate the number of sampling points used to calculate the distance field.}
\label{fig:graph1}
\end{figure}

\begin{figure}[t]
\captionsetup[subfigure]{justification=centering}
\centering
\subcaptionbox{Base network 1 \\ 512-256-128 \label{fig:40-1}}{\includegraphics[width=.32\linewidth]{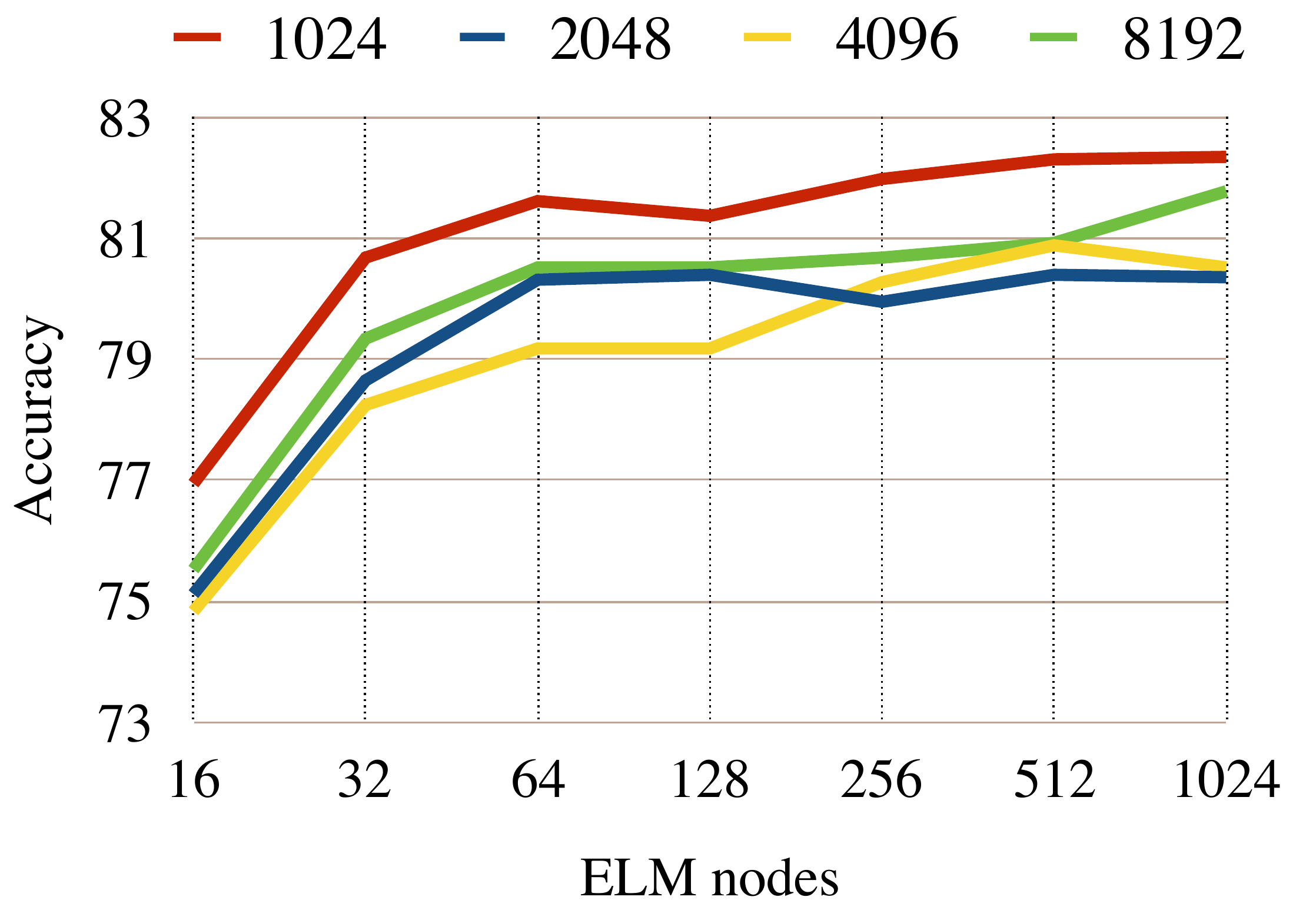}}
\centering
\subcaptionbox{Network 2  \\ 512-512-256-128 \label{fig:40-2}}{\includegraphics[width=.32\linewidth]{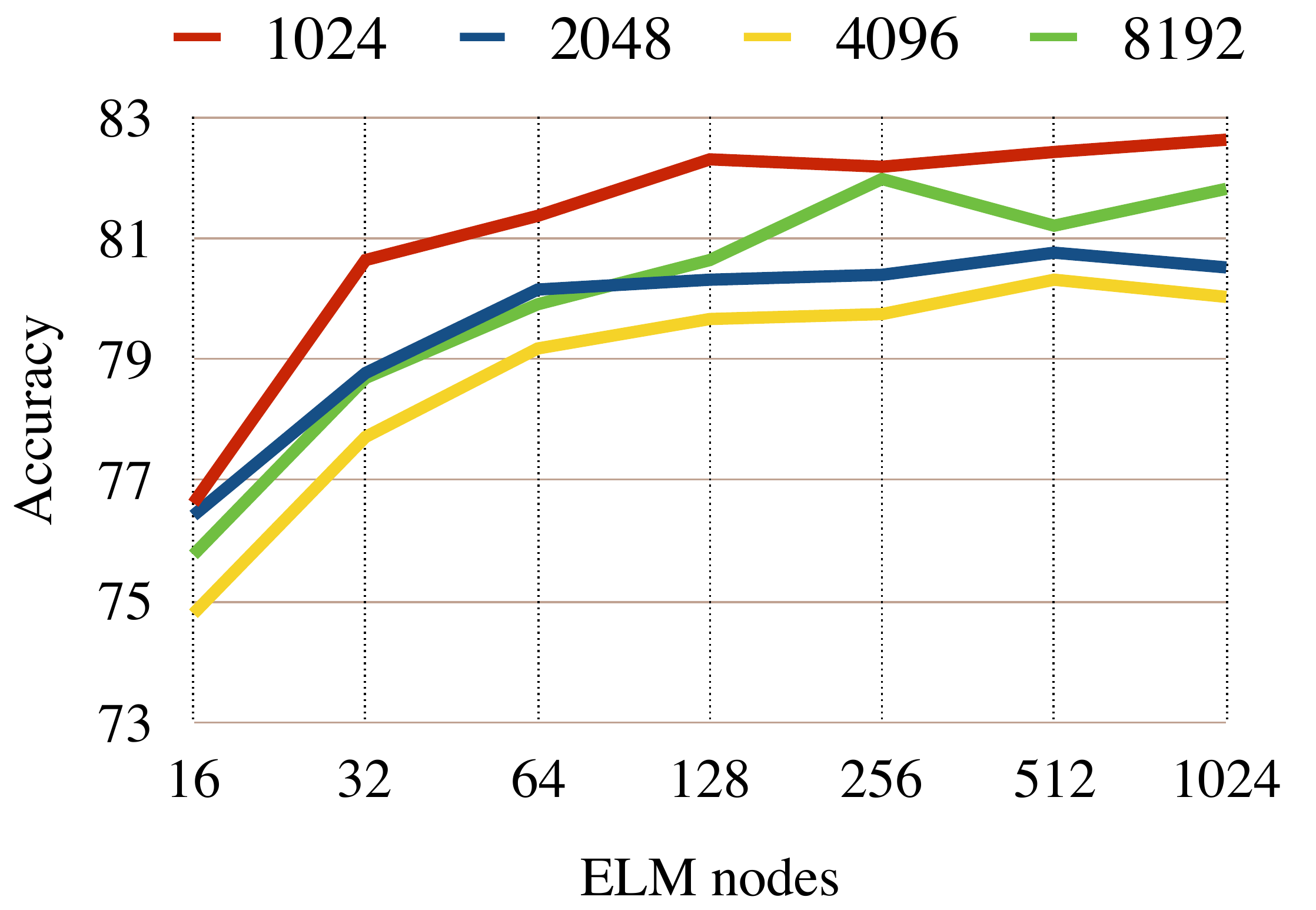}}
\centering
\subcaptionbox{Network 3 \\  512-512-512-256-128 \label{fig:40-3}}{\includegraphics[width=.32\linewidth]{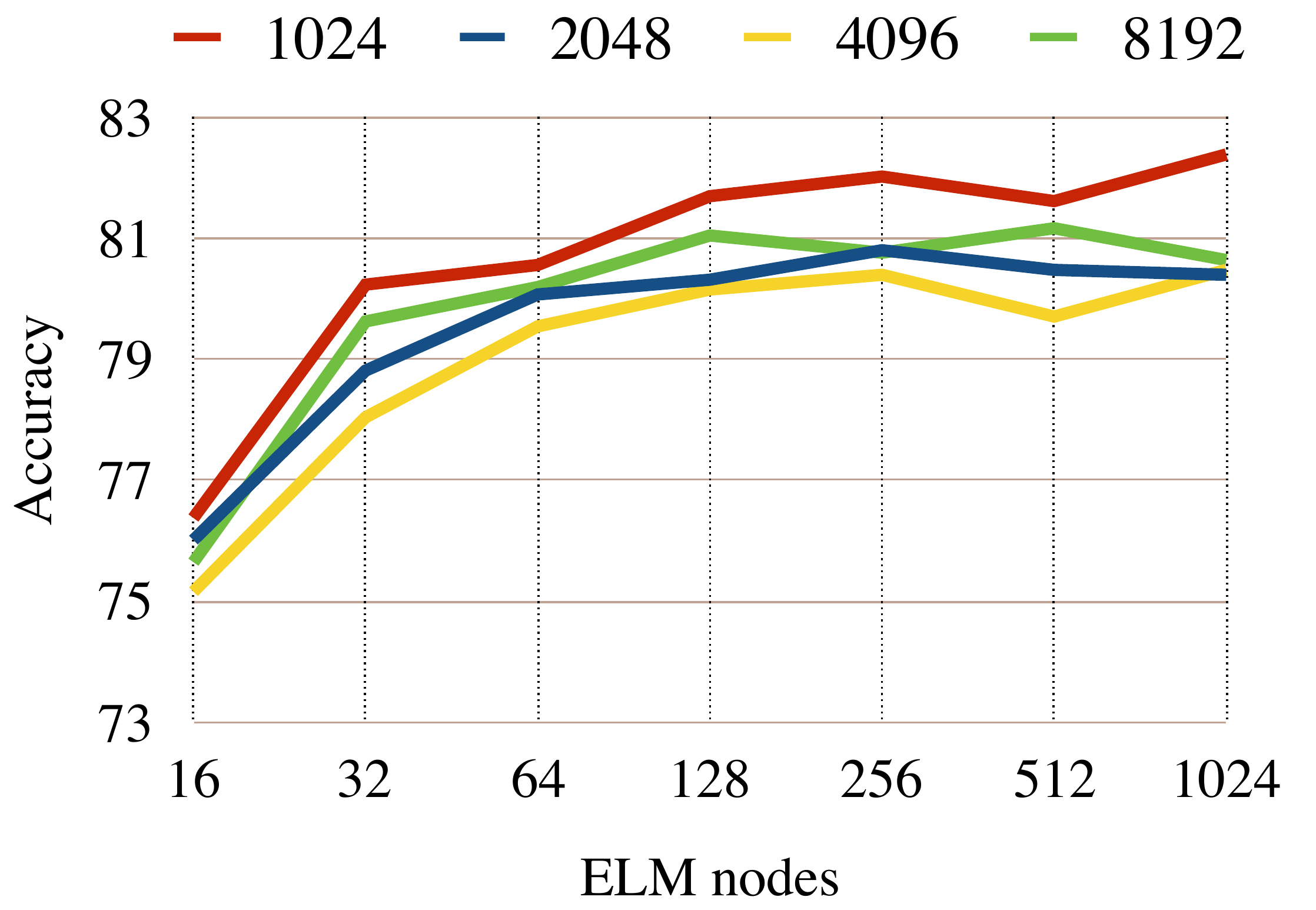}}
   \caption{Accuracy of the proposed method for ModelNet 40 at different sampling densities.}
\label{fig:graph2}
\end{figure}

The results for ModelNet 10 and 40 are shown in figs.~\ref{fig:graph1} and \ref{fig:graph2}, respectively. 

From the results in ModelNet 10, we can observe that as the number of ELM nodes increases, the accuracy also improves. With the proposed representation, only $32$ nodes are required to classify shapes in the ModelNet 10 at $88\%$ accuracy (This value improves to $90.7\%$ when dropout is applied). This indicates that a shape of $n$ surface points can be described by a very compact $32$-dimensional vector, regardless of its initial pose. 

As for sampling density, the greater number of sampling points led to weaker results. This can be understood from the fact that distance field at locations far from the original surface tend to smooth out regardless of the shape of the surface. Therefore, increasing the number of sampling points leads to increasing the chance of sampling regions that are far from the surface. The ELM treats all the sampling points fairly and attempts to encapsulate the entire sampled distance field, thus wasting some descriptiveness on regions that are irrelevant to the actual classification.

Surprisingly, a neural network with just 3 hidden layers was enough to achieve the above results. Increasing the number of layers in the classification neural network resulted in a slight improvement. This demonstrates the powerfulness of our representation, as information required to distinguish the classes of shapes are compactly encapsulated within a small fixed-size vector. 

  
In the case of ModelNet 40, similar trends as ModelNet 10 were observed. The accuracy grew with increased number of nodes, peaking at around $256 - 512$ nodes. The results improved with less number of sampling points, again demonstrating that not all the samplings points contain information that is crucial to the classification accuracy. 

We then took the combinations of the best results and modified other elements, such as adding dropout and modifying the number of neurons and hidden layers in the classification neural network. 

 We compared the results from our method with those from the recent methods. We stress the fact that no pre-training using external datasets nor data augmentation had been conducted to achieve the resulting accuracy. We note that the image-based methods, at the top of the table, rely on pretrained models obtained from other datasets, such as ILSVRC dataset~\cite{russakovsky2015imagenet}. We mainly compare our results with those that only utilize information obtained from point clouds to make the comparison fair.  
 
As can be seen from the results in Table \ref{tab:res}, our method was able to perform better than some of the methods with just a single representation per instance. While most methods rely also on deep neural networks, our classification network is very shallow, requiring short time for convergence as well. As a result, we have significantly reduced the time required for training. We calculated the computational time on a Core-i7-6859K CPU (3.60GHz) computer with $64$GB memory and NVIDIA Quadro P6000 graphics card. For the training time, the fastest of the baselines, PointNet, required $11640$ seconds, while our method, with $4096$ sampling points, $256$ ELM nodes, and base network with hidden layers $512-256-128$, completed training in
 $1696$ seconds (DistField+SVD: $1070$, ELM: $166$, MLP: $460$). However, as our method needs to train ELMs for test data as well, testing time was $0.122$ seconds (DF+SVD: $0.11$, ELM: $0.01$, MLP: $0.002$) per instance, compared to $0.010$ seconds per instance for PointNet.  There are $9840$ training and $2468$ testing instances in ModelNet 40 dataset.  Our process can be made faster by parallelizing the SVD and ELM training, which we have not conducted here.

\subsubsection{Robustness}
We then demonstrate the robustness without data augmentation. The accuracy of the single-view version of MVCNN was $83.0\%$, ours achieved  ${\bf84.9\%}$ with a single representation. This demonstrates that our method is capable of handling rotation even without data augmentation. This  rotation invariance is achieved by the projection of distance field to the canonical space, as we have described in the proposal.  

We then demonstrate the robustness to scale change. When test data was scaled by $0.5$, the accuracy of PointNet lowered to $81.8\%$. Ours achieved $ {\bf83.9\%}$ with the same setting. As proven in the proposal, our carefully designed ELM achieved scale invariance both theoretically and in practice. 

\subsection{Improvement through Subsampling}


To make the classification more robust, we propose to use a  distance field to embed into ELM. To conduct this, we subsample some sampling points from the original set. We use this subset of the sampling points and the corresponding distance values as the input and output of the ELM. Repeating this several times will make the model more robust, as the parameters of the ELM would slightly differ depending on the location of the sampling points. 

 We used a similar setup as the previous experiment. We set $4096$ sampling points around the object point cloud and used a neural network consisting of hidden layers with $512, 256, 128$ nodes. We then sampled $512$ points from the original $2048$ surface points. At each subsampling phase, randomly selected surface points are used to calculate the distance fields. 
 
 
 
%

 \begin{table}[t]
\begin{center}
\begin{tabular}{l|c|c|c|c}

Method  & pretraining & data/instance & ModelNet10  &  ModelNet 40 \\
\hline
\hline
MVCNN~\cite{su2015}&  & multiple & - & 90.1\%\\
Dominant Set~\cite{wang2017dominant}& yes & multiple & - & 93.8\% \\
RotationNet~\cite{kanezaki18}&  & multiple & 98.46\% & 97.37\% \\
\hline
Voxel~\cite{wu2015} & & multiple & 83.5\% &77\%\\ 
GIFT~\cite{bai2016gift} & &multiple& 91.12\%& 83.1\%\\
Authalic~\cite{sinha2016deep} & &multiple& 88.4\%& 83.9\%\\
Pointnet~\cite{qi2016pointnet} & no  &multiple & - &89.2\%\\ 
DeepSet~\cite{zaheer2017deep} & &multiple& - & 90.3\%\\
PANORAMA-NN~\cite{sfikas2017exploiting}& &multiple& 91.1\% & 90.7\%\\
Kdtree~\cite{klokov2017escape}& & multiple&94.0\% & 91.8\% \\
Pointnet++~\cite{qi2017pointnet++} &   &multiple & - &91.9\%\\ 

\hline
Ours & no &1 & 91.9\% & 84.9\%\\
Ours & no &16 & 92.7\% & 85.8\%\\
\hline
\end{tabular}
\end{center}
\caption{Comparison of results on Modelnet $10$ and $40$. Our method performs better than some of the state of the art methods with just one representation per instance. The results improved with sampling subset of original point clouds.}
\label{tab:res}
\end{table}

 The bottom row of Table~\ref{tab:res} shows the best results after data subsampling. As can be seen from the numbers, the results improved. However, the margin of improvement is small. This is the consequence of the robustness of our representation demonstrated in the first experiment. Despite the fact that a new distance field is created from a subset of surface points, the representation is very robust and is only slightly different from the original.

\section{Conclusion}

Our method derives from the belief that shapes should be preprocessed into a unified parametric space rather than trying to prepare every possible viewpoint of the shapes manually through data augmentation. We arrived at the idea from the observation that if neural networks can learn a function that separates various classes accurately, they can also be used to learn a function that represents one particular shape and use the network itself as a feature representing the shape.

The experimental results demonstrated the validity of our idea. Our method is not only useful for achieving various invariances that made representation and classification of unstructured point difficult, but also effective in the actual classification task. The resulting representation is compact and powerful, that even a single representation from our method can achieve an accuracy close to the state of the art. Note that no end-to-end training is conducted, and that only a shallow neural network is required to achieve all of this. By changing the surface sampling points, we managed to conduct data augmentation, which led to even higher accuracy. 

The results can be further improved, as the random bases used in ELM are not tuned to work positively towards higher classification accuracy. As future work, we will pursue a method to tune the bases in the ELM to improve the classification accuracy and also make the weights even more compact. 


\clearpage

\bibliographystyle{splncs}
\bibliography{egbib}

\begin{thebibliography}{10}

\bibitem{wu2015}
Wu, Z., Song, S., Khosla, A., Yu, F., Zhang, L., Tang, X., Xiao, J.:
\newblock 3d shapenets: A deep representation for volumetric shapes.
\newblock In: Proceedings of the IEEE Conference on Computer Vision and Pattern
  Recognition. (2015)  1912--1920

\bibitem{su2015}
Su, H., Maji, S., Kalogerakis, E., Learned-Miller, E.:
\newblock Multi-view convolutional neural networks for 3d shape recognition.
\newblock In: Proceedings of the IEEE international conference on computer
  vision. (2015)  945--953

\bibitem{sinha2016deep}
Sinha, A., Bai, J., Ramani, K.:
\newblock Deep learning 3d shape surfaces using geometry images.
\newblock In: European Conference on Computer Vision. (2016)  223--240

\bibitem{huang2006extreme}
Huang, G.B., Zhu, Q.Y., Siew, C.K.:
\newblock Extreme learning machine: theory and applications.
\newblock Neurocomputing \textbf{70}(1) (2006)  489--501

\bibitem{qi2016pointnet}
Qi, C.R., Su, H., Mo, K., Guibas, L.J.:
\newblock Pointnet: Deep learning on point sets for 3d classification and
  segmentation.
\newblock In: Proceedings of the IEEE Conference on Computer Vision and Pattern
  Recognition. (2016)

\bibitem{tangelder2004}
Tangelder, J.W., Veltkamp, R.C.:
\newblock A survey of content based 3d shape retrieval methods.
\newblock In: Shape Modeling Applications, 2004. (2004)  145--156

\bibitem{song2015}
Song, S., Lichtenberg, S.P., Xiao, J.:
\newblock Sun rgb-d: A rgb-d scene understanding benchmark suite.
\newblock In: Proceedings of the IEEE conference on computer vision and pattern
  recognition. (2015)  567--576

\bibitem{chang2015}
Chang, A.X., Funkhouser, T., Guibas, L., Hanrahan, P., Huang, Q., Li, Z.,
  Savarese, S., Savva, M., Song, S., Su, H.,  et~al.:
\newblock Shapenet: An information-rich 3d model repository.
\newblock arXiv preprint arXiv:1512.03012 (2015)

\bibitem{maturana2015}
Maturana, D., Scherer, S.:
\newblock Voxnet: A 3d convolutional neural network for real-time object
  recognition.
\newblock In: Intelligent Robots and Systems (IROS), 2015 IEEE/RSJ
  International Conference on, IEEE (2015)  922--928

\bibitem{wu2016learning}
Wu, J., Zhang, C., Xue, T., Freeman, B., Tenenbaum, J.:
\newblock Learning a probabilistic latent space of object shapes via 3d
  generative-adversarial modeling.
\newblock In: Advances in Neural Information Processing Systems. (2016)  82--90

\bibitem{sedaghat2017orientation}
Sedaghat, N., Zolfaghari, M., Amiri, E., Brox, T.:
\newblock Orientation-boosted voxel nets for 3d object recognition.
\newblock In: British Machine Vision Conference (BMVC). (2017)

\bibitem{Wang2017}
Wang, C., Pelillo, M., Siddiqi, K.:
\newblock Dominant set clustering and pooling for multi-view 3d object
  recognition.
\newblock In: Proceedings of the British machine vision conference. (2017)

\bibitem{qi2016volumetric}
Qi, C.R., Su, H., Nie{\ss}ner, M., Dai, A., Yan, M., Guibas, L.J.:
\newblock Volumetric and multi-view cnns for object classification on 3d data.
\newblock In: Proceedings of the IEEE Conference on Computer Vision and Pattern
  Recognition. (2016)  5648--5656

\bibitem{shi2015deeppano}
Shi, B., Bai, S., Zhou, Z., Bai, X.:
\newblock Deeppano: Deep panoramic representation for 3-d shape recognition.
\newblock IEEE Signal Processing Letters \textbf{22}(12) (2015)  2339--2343

\bibitem{qi2017pointnet++}
Qi, C.R., Yi, L., Su, H., Guibas, L.J.:
\newblock Pointnet++: Deep hierarchical feature learning on point sets in a
  metric space.
\newblock In: Advances in Neural Information Processing Systems. (2017)
  5105--5114

\bibitem{klokov2017escape}
Klokov, R., Lempitsky, V.:
\newblock Escape from cells: Deep kd-networks for the recognition of 3d point
  cloud models.
\newblock In: Proceedings of the International Conference on Computer Vision.
  (2017)

\bibitem{soltani2017synthesizing}
Soltani, A.A., Huang, H., Wu, J., Kulkarni, T.D., Tenenbaum, J.B.:
\newblock Synthesizing 3d shapes via modeling multi-view depth maps and
  silhouettes with deep generative networks.
\newblock In: Proceedings of the IEEE Conference on Computer Vision and Pattern
  Recognition. (2017)  1511--1519

\bibitem{zanuttigh2017deep}
Zanuttigh, P., Minto, L.:
\newblock Deep learning for 3d shape classification from multiple depth maps.
\newblock In: Proceedings of IEEE International Conference on Image Processing
  (ICIP). (2017)

\bibitem{nair2010rectified}
Nair, V., Hinton, G.E.:
\newblock Rectified linear units improve restricted boltzmann machines.
\newblock In: Proceedings of the 27th international conference on machine
  learning (ICML-10). (2010)  807--814

\bibitem{maas2013rectifier}
Maas, A.L., Hannun, A.Y., Ng, A.Y.:
\newblock Rectifier nonlinearities improve neural network acoustic models.
\newblock In: in ICML Workshop on Deep Learning for Audio, Speech and Language
  Processing. (2013)

\bibitem{turk1994zippered}
Turk, G., Levoy, M.:
\newblock Zippered polygon meshes from range images.
\newblock In: Proceedings of the 21st annual conference on Computer graphics
  and interactive techniques. (1994)  311--318

\bibitem{tang2016extreme}
Tang, J., Deng, C., Huang, G.B.:
\newblock Extreme learning machine for multilayer perceptron.
\newblock IEEE transactions on neural networks and learning systems
  \textbf{27}(4) (2016)  809--821

\bibitem{russakovsky2015imagenet}
Russakovsky, O., Deng, J., Su, H., Krause, J., Satheesh, S., Ma, S., Huang, Z.,
  Karpathy, A., Khosla, A., Bernstein, M.,  et~al.:
\newblock Imagenet large scale visual recognition challenge.
\newblock International Journal of Computer Vision \textbf{115}(3) (2015)
  211--252

\bibitem{wang2017dominant}
Wang, C., Pelillo, M., Siddiqi, K.:
\newblock Dominant set clustering and pooling for multi-view 3d object
  recognition.
\newblock In: Proceedings of British Machine Vision Conference (BMVC). (2017)

\bibitem{kanezaki18}
Kanezaki, A., Matsushita, Y., Nishidai, Y.:
\newblock Rotationnet: Learning object classification using unsupervised
  viewpoint estimation.
\newblock In: Computer Vision and Pattern Recognition (CVPR), 2018 IEEE
  Conference on. (2018)

\bibitem{bai2016gift}
Bai, S., Bai, X., Zhou, Z., Zhang, Z., Latecki, L.J.:
\newblock Gift: A real-time and scalable 3d shape search engine.
\newblock In: Computer Vision and Pattern Recognition (CVPR), 2016 IEEE
  Conference on. (2016)  5023--5032

\bibitem{zaheer2017deep}
Zaheer, M., Kottur, S., Ravanbakhsh, S., Poczos, B., Salakhutdinov, R.R.,
  Smola, A.J.:
\newblock Deep sets.
\newblock In: Advances in Neural Information Processing Systems. (2017)
  3394--3404

\bibitem{sfikas2017exploiting}
Sfikas, K., Theoharis, T., Pratikakis, I.:
\newblock Exploiting the panorama representation for convolutional neural
  network classification and retrieval.
\newblock In: Eurographics Workshop on 3D Object Retrieval. (2017)

\end{thebibliography}
\end{document}